\documentclass[sigconf]{acmart}
\usepackage{booktabs} 
\usepackage{makecell}
\usepackage{multirow}

\copyrightyear{2019} 
\acmYear{2019} 
\setcopyright{acmlicensed}
\acmConference[FDG '19]{The Fourteenth International Conference on the Foundations of Digital Games}{August 26--30, 2019}{San Luis Obispo, CA, USA}
\acmBooktitle{The Fourteenth International Conference on the Foundations of Digital Games (FDG '19), August 26--30, 2019, San Luis Obispo, CA, USA}
\acmPrice{15.00}
\acmDOI{10.1145/3337722.3337755}
\acmISBN{978-1-4503-7217-6/19/08}

\acmArticle{1}
\acmPrice{15.00}
\begin{document}
\title{Making CNNs for Video Parsing Accessible}
\subtitle{Event Extraction from \textit{DOTA2} Gameplay Video using Transfer, Zero-Shot, and Network Pruning}

\author{Zijin Luo}
\affiliation{\institution{Georgia Institute of Technology}}
\email{zijinluo@gatech.edu}

\author{Matthew Guzdial}
\affiliation{\institution{Georgia Institute of Technology}}
\email{mguzdial3@gatech.edu}

\author{Mark Riedl}
\affiliation{\institution{Georgia Institute of Technology}}
\email{riedl@cc.gatech.edu}

\begin{abstract}
The ability to extract sequences of game events for high-resolution e-sport games has traditionally required access to the game's engine. 
This serves as a barrier to groups who don't possess this access.
It is possible to apply deep learning to derive these logs from gameplay video, but it requires computational power that serves as an additional barrier.
These groups would benefit from access to these logs, such as small e-sport tournament organizers who could better visualize gameplay to inform both audience and commentators.
In this paper we present a combined solution to reduce the required computational resources and time to apply a convolutional neural network (CNN) to extract events from e-sport gameplay videos. 
This solution consists of techniques to train a CNN faster and methods to execute predictions more quickly.
This expands the types of machines capable of training and running these models, which in turn extends access to extracting game logs with this approach.
We evaluate the approaches in the domain of \textit{DOTA2}, one of the most popular e-sports. 
Our results demonstrate our approach outperforms standard backpropagation baselines.
\end{abstract}

%


\maketitle


\section{Introduction}
Game logs describe gameplay as sequences of events that represents player experience.
A game log is a record of game events, which represent the major actions affecting the progress and direction of the game. 
Game developers collect game logs in order to better understand how players interact with their game and to find balance changes and bug fixes. 
However, game log access is typically restricted to only a game's developers, as  they are generated by a game's backend or on a studio's servers.
Access to game logs can benefit many communities in and around games. 
For example in e-sports, without game studio support game logs could benefit viewers, commentators, players and coaches through summarizing information and replays. 
In broad games academia, access to game logs can mean greater understanding and analysis of how people play games without public game logging systems.

Machine learning algorithms, specifically convolutional neural networks (CNNs), are widely utilized to perform video action recognition from video \cite{soomro2012ucf101}. 
One could imagine applying a CNN to the problem of deriving game logs, sequences of gameplay events, from video.
However, to run a deep CNN requires significant computational resources and the time to both train and execute a model.
Moreover, high-resolution objects and complex textures, like those found in modern games, increase the difficulty of this problem and the associated computational resources to solve it.
This limitation restricts the usage of CNNs to only those with the high-performance machines, especially in cases like a e-sport tournament livestream.
In such a case the CNN game action recognition model would need to run in near real-time in order to keep up with the live broadcast so its output could be used to inform casters and the audience. 
These types of tournaments would also likely lack the kinds of high-powered machines one typically associates with deep learning, and one would want any such machines to go to actually playing the e-sport.
Therefore, even outside of the accuracy of such a model, the computation power requirements and execution time of this model would matter hugely in determining the success of this approach.

In this paper, we present approaches to derive accessible CNN models.
"Accessible" for the purposes of this paper referring to hardware accessibility only, however we note that hardware accessibility is tied to issues like financial and technical accessibility.
Therefore we focus on CNN models that can be obtained and deployed with little computational resources and time, and can convert from gameplay video to a game log-like sequence of game events in real time.
Our goal is to find solutions to reduce requirements for obtaining and executing deep CNNs on low-performance machines.
For our domain we employ \textit{Defense of the Ancient 2} or \textit{DOTA2} due to the fact that it is a popular e-sport and has stylized visuals (meaning that one cannot simply apply real life event detection models).
The remainder of this paper is organized as follows: First, we cover the associated work in recognizing and predicting the actions or activities from gameplay videos. 
Second, we introduce the standard CNN approach, trained on a target dataset of \textit{DOTA2} game video and associated events and actions directly.
Third, we present experiments to decrease the required computational power to train and run these models, comparing each to standard approach baselines. 
Our primary contributions are the experiments and evaluations that prove one can reduce the amount of computational power and time to train and run a CNN for event extraction from gameplay video.
We make available a public dataset of player actions for the e-sports game \textit{DOTA2} as a secondary contribution in this paper. \footnote{https://github.com/icpm/dota2-dataset}


\section{Related Work}
Player modeling \cite{yannakakis2013player}, the study of computationally modeling the players' actions in video games, represents a related domain to this paper, given that it requires a representation of player actions as input.
Nonetheless, most player modeling systems and methods are typically developed upon the necessary assumption of sufficient computational power and/or access to comprehensive data of game events. 
Player modeling research in the domain of commercial games has traditionally required support or partnership with existing game developers \cite{drachen2009player, sabik2015data,makarovych2018like}. 
As an alternative, some prior player modeling researches explored the possibility of creating clones of past commercial games \cite{togelius20102009,shaker2015towards,siu2017evaluating}, but this requires both design and game development knowledge, takes a large amount of time, and is not tenable for modern, large-scale games.
\citeauthor{camilleri2017towards} \cite{camilleri2017towards} proposed to construct a general model of player affect across distinct games to make player modeling accessible to more people. However this approach still requires access to the logging system of each game.

Prior approaches exist to derive approximated game logs from gameplay video. However, the majority of these approaches rely on hand-authored event definitions \cite{jacob2014non} or a secondary machine vision library like OpenCV \cite{bradski2000opencv} and access to all of the images in the game \cite{guzdial2016game,bao2017extracting}. This latter approach is only viable for two-dimensional games with limited sets of possible images for each game component (called a ``sprite'').

\begin{figure*}
	\includegraphics[width=\linewidth]{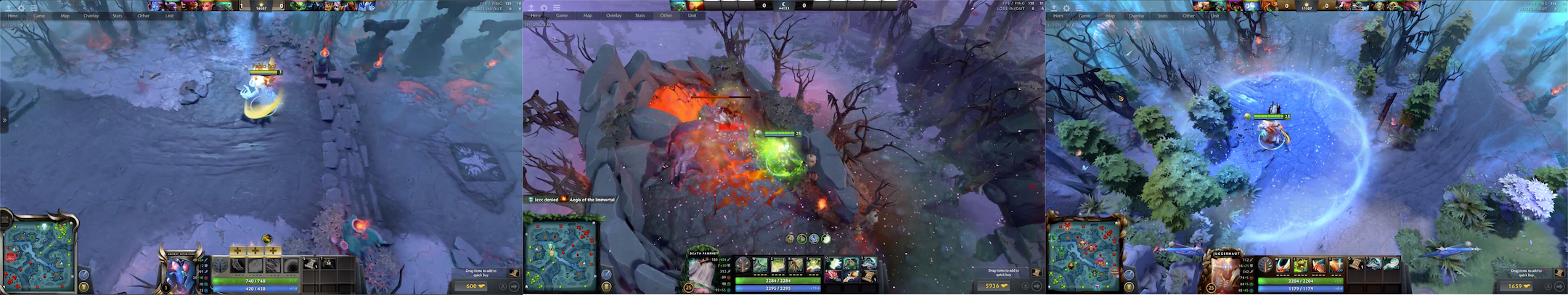}
	\centering
	\caption{Screenshot of three of the ten categories of game events: using Black King Bar (left), fighting Roshan (middle), and using Shiva's Guard (right).}
	\label{fig:DOTA2Screenshot}
\end{figure*}

\citeauthor{mccallum2000maximum} \cite{mccallum2000maximum} proposed to use the Maximum Entropy Markov Models to perform information extraction from images inputs, but we choose to focus on the deep learning approaches, which have shown great success.
Deep neural networks have previously been applied to predict or approximate player experience in a summarizing capacity.
For example, approaches applying CNNs to directly predict player experience metrics like fun and challenge from levels \cite{guzdial2016deep} or game logs \cite{liao2017deep}.
\citeauthor{summerville2016learning} \cite{summerville2016learning} derived logs of player movement from gameplay video with OpenCV and then applied a long short-term memory (LSTM) recurrent neural network to predict player paths through new levels. However, their approach only focused on player movement events.
\citeauthor{karavolos2018using} \cite{karavolos2018using} utilized CNNs as a surrogate model of gameplay in order to predict game level balance based on automated high-level summarizing metrics. 
\citeauthor{fulda18threat} \cite{fulda18threat} predicted player actions from Skyrim video based on existing machine vision models, but presented inconclusive results.
\citeauthor{luo2018player} \cite{luo2018player}, the work most similar to ours, applied CNNs to convert from gameplay video to representations of player experience for a Super Mario clone, Mega Man, and Skyrim through transfer learning. Our work differs from this prior work given that we focus on decreasing computation power and time costs for training and running these approaches, and that we focus on a multiplayer, modern e-sport in \textit{DOTA2}.

There exist applications of activity recognition for real-world games and sports \cite{zhu2006player}. However, this is largely enabled by the existence of large datasets of real-world human activity \cite{soomro2012ucf101}. Given many games, like our chosen domain of \textit{DOTA2}, have distinct, unrealistic aesthetics, we would not anticipate these approaches to transfer without modification.

Existing prior approaches have improved CNN training time significantly by utilizing explicit input normalization and decreasing the variance.
\citeauthor{ioffe2015batch} \cite{ioffe2015batch} introduced the use of batch normalization to minimize changes over each iteration during the training process, which is notorious for slowing down the training by requiring lower learning rates and parameter initialization. 
Weight normalization \cite{salimans2016weight} is also a solution to simplify the optimization problem and speed up the convergence of stochastic gradient descent during the training process.
Another particular method to accelerate training is to use special hardware \cite{ovtcharov2015accelerating}. However, this approach demands additional optimization and support from hardware manufacturers. 
Those techniques are either general methods to accelerate training or require specialized hardware. 
They are not designed toward real world action recognition domains, not for games, and they do not take into account accessibility concerns. 
We demonstrate that one can use other techniques to accelerate the training process much faster without additional computational power or specialized hardware.

There exists some approaches to pruning a deep neural network to reduce the parameters of the learned model \cite{han2015deep}.
These pruning methods such as network slimming \cite{liu2017learning} are typically associated with large, redundant CNN architectures like VGG-19 \cite{simonyan2014very}.
These approaches would not be appropriate to ResNet series models, like the one we employ in this paper.

\section{Standard Method}

We begin by describing what one might consider the standard application of convolutional neural networks to the problem of event recognition from gameplay video in \textit{DOTA2}. This will then serve as a baseline by which we compare our attempts to improve training and execution time, and to minimize computational power requirements. We refer to this baseline as the ``Standard Method".

Our standard method uses ResNet152 \cite{he2016deep} as the architecture for our convolutional neural network. 
We chose ResNet152 as it represents a state of the art model on many computer vision problems.
Despite this paper's focus on accessibility, high accuracy is still a necessary condition for any implementation that attempts to derive sequences of game events from gameplay video.
ResNet152 also benefits from the fact that it can still run on unspecialized machines with only a single graphical processing unit (GPU) like those found on most gaming computers. 
This type of computer serves as our hardware goal for the purposes of this paper, given our motivating example of small-scale e-sport tournaments. 
However, we note that while ResNet152 can run on machines with a single GPU, it still has a number of drawbacks. Notably, the training time can still be lengthy, the execution time can be well below real-time, and the computational requirements (particularly memory) can still make this model inaccessible to such machines for our stated purpose.
We chose to use \textit{Double Cross Entropy} as the loss function, and \textit{Adam} as the optimizer.

In this paper, we focus on a ResNet152 model to predict a one-hot encoding of game events from a single gameplay frame.
This means that for each frame of input the model predicts a single vector, with each index of the final vector representing an in-game event.
We take the maximally activated index as the final prediction of the model, of what event is happening in the particular input game.
For the purposes of this paper, we do not have any null gameplay video frames, in which no in-game events occur.
However, this can be incorporated into a model by introducing a threshold of minimal activation under which no events are predicted \cite{luo2018player}.

Our model setup allows one to either take a final trained model and predict what is currently happening in a streamed match or to parse an existing, recorded gameplay video. 
The former of these would require that the system execute in real-time, giving casters and/or audience members a sense of what is currently happening.
The latter of these would lead to a sequence of game events that approximates the true game log, depending on the performance of the trained model.
Note that this model is not limited to one-hot encoding, one can turn it into multi-hot encoding by changing the loss function and event representation.
However, we felt that for an initial study a one-hot representation would be more appropriate for explanatory purposes and to determine if this approach is even tenable in a hardware accessibility context.

We chose to make use of only the ten most recognizable game events from \textit{DOTA2} in this paper as chosen by a \textit{DOTA2} expert and author on this paper. 
We made this choice because we desire easily interpretable output from our model, to both expert and casual fans.
Further, we focused on only these most important events given that this is an initial investigation into the appropriateness of these approaches in the domain of \textit{DOTA2}.
Given that ResNet style models have shown success at datasets with thousands of classes, there is no reason to suspect that our results are only applicable to problems with few classes. Our ten chosen event types are: using Black King Bar, using Eul's Scepter of Divinity, using a Glyph, Ending the game, Roshan fight, using Shiva's Guard, activating a Shrine, team fight, teleport, and tower destruction.
We do not describe these events in detail here, but we note that they are a mix of important events involving multiple characters and individual characters employing powerful items or abilities.
These events are notably independent of player character class in \textit{DOTA2}, so what particular characters are involved should not affect the classifier's performance.
Beyond the importance of each of these events they all also have unique visual effects, which is an important factor in the success of CNN action recognition approaches.
See examples of three of these events in Figure \ref{fig:DOTA2Screenshot}.

To train any CNN model, one requires a dataset. 
We collected a dataset of 10 three second gameplay clips of each event.
We extracted frames from these videos at 30 frames per second (FPS), as the collected videos were recorded with variable FPS.
This approach can deal with any FPS given that the ResNet152 model simply takes in individual frames.
This lead to a total dataset size of 9,000 frames, each tagged with a single gameplay event, which we represented in the one-hot vector representation described above.

To evaluate this approach, we utilize the same \textit{DOTA2} dataset with an 80-20 train-test split, leaving us with 7,200 training frames and 1,800 test frames.
We uniformly drew from the dataset to collect the test split, in order to ensure that any variation between the approaches was caused by variations in our approach and not variations in the data.
We acknowledge that splitting consecutive and similar frames in both training and testing datasets might end up artificially increasing accuracy. 
However, we use the same training and test splits across all approaches, meaning that the comparative accuracy increases are independent of this potential issue.

We note that exact accuracy does not matter given that we are comparing between the approaches, not looking to validate the use of CNNs to extract game logs generally, which has already shown to have impressive results \cite{luo2018player}.
The primary evaluation directions we chose were the training speed and executing speed. For the training speed, we recorded the training loss, training accuracy, testing loss, and testing accuracy for future usage and comparison. 
For the executing speed, we stored the number of parameters in the model and the time required to run the whole \textit{DOTA2} dataset. 
We include detailed comparisons in the following sections, but note that the standard approach already performed well with a final test accuracy of roughly 95\% after convergence. 
This supports our choice of ResNet152, given the importance of accuracy in this context. 
However, there is still room for improvement, as a 95\% accuracy rate would still lead to approximately three mistakes every second when parsing livestream data at 60 FPS.


\section{Decreasing Training Time with Transfer Learning}
The training of a CNN classifier is both the major determinant of a classifier's performance and the most time-consuming part of applying a CNN.
The exact amount of time required depends upon the hardware on which the CNN is trained, multiple GPUs allow for greater parallelism, which in turn decreases training time.  
Given that most computers built for gaming have a single GPU, training time represents a serious boundary to accessibility for those with this hardware.
To train a large-scale model like ResNet152 is additionally a long and resource intensive task on such a computer.
Given sufficient time one could train a ResNet152 model on a computer with a single GPU, but this assumes the ability to dedicate the computer entirely to the training process for hours (depending on the size of the dataset and complexity of the problem).

We apply two different transfer learning approaches to decrease training time: teacher-student \cite{wong2016sequence,furlanello2018born} and neural style transfer \cite{gatys2016image}. 
The intuition behind both approaches is that retraining an existing ResNet152 model trained on another dataset will be faster than training a new ResNet152 model from scratch. 

In this case we drew upon ImageNet \cite{deng2009imagenet} as the existing dataset to transfer knowledge from.
There are two primary reasons for choosing ImageNet. 
First, ImageNet is a massive dataset, containing about 80,000 hierarchical classes with an average of 500-1000 images and is currently applied to test state-of-the-art approaches.
Second, ImageNet pre-trained models, including ResNet152 models, are publicly available in the Pytorch Model Zoo.
The first of these points ensures that this dataset is of suitable quality to lead to an appropriate final model, and the second point ensures that the required items to run these approaches are publicly available.

\subsection{Teacher-Student Approach}

In this section, we discuss an approach to use teacher-student transfer learning, sometimes called \textit{knowledge distillation} \cite{wong2016sequence, furlanello2018born}. 
The teacher-student method replicates the weights of the pre-trained ``teacher" model to a new, generally smaller ``student" model. 
The intuition behind the teacher-student method is that the student model can leverage and modify the features learned by the teacher model on a new task.
The advantage of this method is we only need to train the teacher model once, and we can apply this ``teacher`` model to many different ``student`` models, which require smaller datasets and training times.
This approach also ensures that the user does not need to personally train the teacher model, if one is otherwise available.
\citeauthor{luo2018player} \cite{luo2018player} also demonstrated that a model pre-trained on ImageNet could successfully serve teacher model for gameplay video student models.

For our case we took a ResNet152 model pre-trained on ImageNet, removed the final classification layer for classifying on the ImageNet classes and added a new layer to predict across our ten possible events. 
We then trained this student model on the 7,200 \textit{DOTA2} training frames from our dataset. 

\subsection{Teacher-Student Approach Evaluation}
\label{sec: ts-eval}

\begin{figure}
	\includegraphics[width=\linewidth]{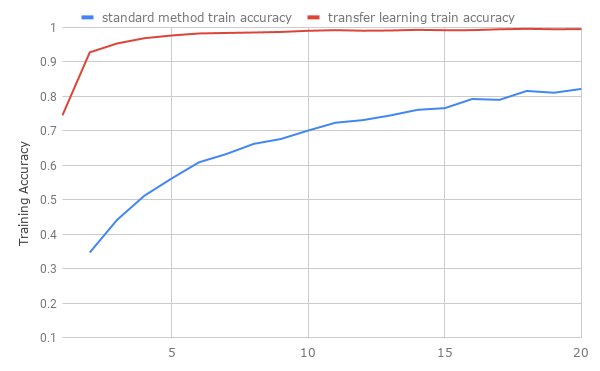}
	\caption{The training accuracy over the first 20 iterations for the teacher-student approach and standard method baseline}
	\label{fig:transfer_train_acc}
\end{figure}

\begin{figure}
	\includegraphics[width=\linewidth]{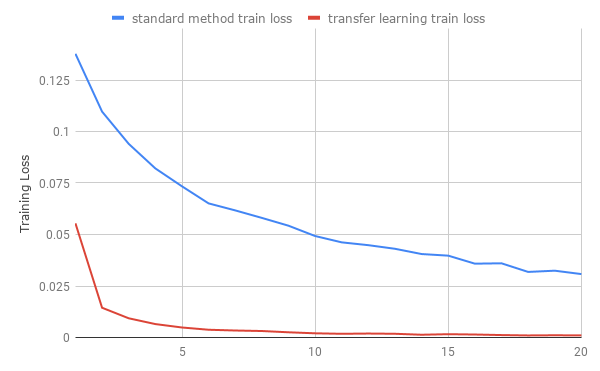}
	\caption{The training loss over the first 20 iterations for the teacher-student approach and standard method baseline}
	\label{fig:transfer_train_loss}
\end{figure}

In this section, we evaluate the performance of the teacher-student approach in terms of required training time and accuracy.
We apply the standard method as a baseline, which trains a ResNet152 model via backpropagation from scratch.
We compare the training accuracy and the loss between the teacher-student approach and the standard approach baseline during the training process. 
Figure \ref{fig:transfer_train_acc} and Figure \ref{fig:transfer_train_loss} represent the training accuracy and loss trend respectively over first 20 iterations for both approaches. 
The red line represents the teacher-student model, and the standard method is in blue.

Both figures demonstrate that the teacher-student approach converges faster in comparison to the standard approach. 
More specifically, the teacher-student model can reach near 100\% training accuracy after five iterations, with the loss approaching zero and remaining steady. 
However, the standard method is still in the converging stage after twenty iterations and both accuracy and loss curves have fluctuations.
Note that the standard method took roughly fifty iterations to fully converge, representing an improvement of roughly an order of magnitude.
In terms of final testing accuracy, the teacher-student approach achieves 99.8\% test accuracy after converging, compared to 94.6\% for the standard method.
While this difference may seem small, over a second in a live stream it would be the difference between an average of no errors and three errors. 
Further, our primary focus is in reducing training time, which teacher-student accomplishes compared to the standard method.

We demonstrate that the teacher-student transfer learning approach is both more efficient and more accurate in comparison to the standard method. 
Based on the empirical data we collected, applying the teacher-student approach only takes 10\% of the required training epochs of the standard method to converge.
This indicates a strong improvement in terms of accessibility for those with time constraints, especially those without high-end hardware.

\subsection{Kernel Neural Style Transfer}
The initial teacher-student transfer learning approach presented excellent results. However, it is possible that the aesthetic differences between the real world images of ImageNet and the stylized aesthetic of \textit{DOTA2} hampered the transfer process. In other words, it is possible the generic features extracted from real world images could have been adapted even more swiftly to the \textit{DOTA2} aesthetic with a different approach.

Focusing on altering aesthetics, we took inspiration from neural style transfer \cite{gatys2016image}. In this approach a source image with one particular style and a target image of a different style are used, producing a new image with the content of the source image and the style of the target image. Naively, one might imagine applying this approach to every image in the ImageNet dataset, somehow choosing a candidate image with \textit{DOTA2}'s aesthetic and then stylizing the original ImageNet image to more reflect \textit{DOTA2}'s style. 
However, this approach would require a major time commitment. 
First, the time to process and stylize the ImageNet dataset and then fully training ResNet152 with this new, stylized dataset.

As an alternative, we instead chose to run style transfer on the learned features of the ResNet152 model trained on ImageNet. 
We call this approach \textbf{Kernel Neural Style Transfer} (KNST).
This approach is inspired by neural style transfer and generative adversarial networks or GANs, to try to accelerate the training process by updating the trained ResNet152 ImageNet classifier kernels with the \textit{DOTA2} style.
We utilized a CNN as a stylizer to select features from a \textit{DOTA2} gameplay image as the target style for the content of the classifier's first kernel.
The motivation is that we would like use style transfer to improve the prediction accuracy while decreasing training time since the neural networks could converge faster if the classifier already has \textit{DOTA2}-like features from the stylizer.
We applied this approach to the very first layer of the CNN given that this layer extracts the general features from the input image, and these features have proven useful in zero-shot learning \cite{chao2016empirical}.
Rather than adding one more layer before the first kernel layer in classifier, KNST utilizes another model to stylize the classifier’s layer. The most fundamental difference between these two approaches is where new features come from, either an internal layer extracts the features from the training input frame as in teacher-student or an external stylizer model extracts the features from an independent picture of this domain. 

\begin{figure}
	\includegraphics[width=\linewidth]{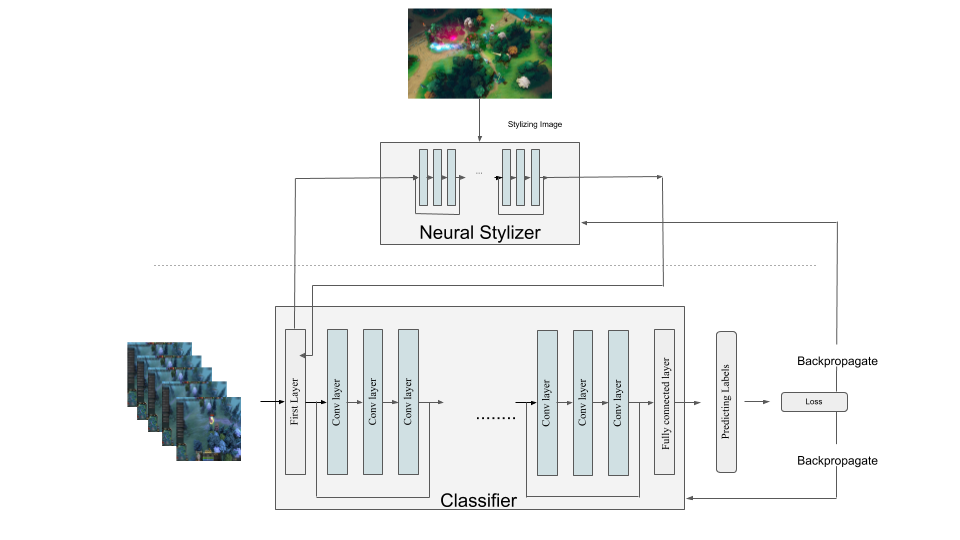}
	\caption{The illustration of Kernel Neural Style Transfer training process.}
	\label{fig:knst_mech}
\end{figure}

Shown in the Figure \ref{fig:knst_mech}, the structure of this combined CNN system is as follows. 
First, we prepared a residual CNN with 5 residual blocks as the ``stylizer`` model, which utilizes convolutional layers to decompose the input kernel and de-convolutional layers to recompose the output kernel. 
Next, we prepare a newly initialized ResNet152 model as the ``classifier`` model that takes in a batch of frames from our dataset as the input and outputs a prediction over the \textit{DOTA2} event types. 
In each training iteration, we feed the first layer of the classifier model to our stylizer model to obtain a stylized kernel and copy the weight of the updated kernel back into the classifier model.
We then feed the frames from our \textit{DOTA2} dataset to our classifier model, essentially doing an epoch of normal training. 
After acquiring the predicted labels, we computed the cross-entropy loss from the predictions and the real labels. 
At this point, we backpropagate this loss value to both the stylizer model and classifier model to update weights.
This procedure guarantees the stylizer model and the classifier model both reflect the training dataset. 

\subsection{Kernel Neural Style Transfer Evaluation}
In the previous section, we introduced the KNST and its implementation.
We need to evaluate if KNST actually helps improve the training time, and how this compares to the teacher-student transfer approach.
In this section, we evaluated KNST using the same method as in Section \ref{sec: ts-eval}.
If we see a similar or greater improvement utilizing KNST as the teacher-student approach in comparison to the standard method that would lend evidence to the importance of aesthetic differences between real world images and games with a \textit{DOTA2}-like visual aesthetic.

\begin{figure}
	\includegraphics[width=\linewidth]{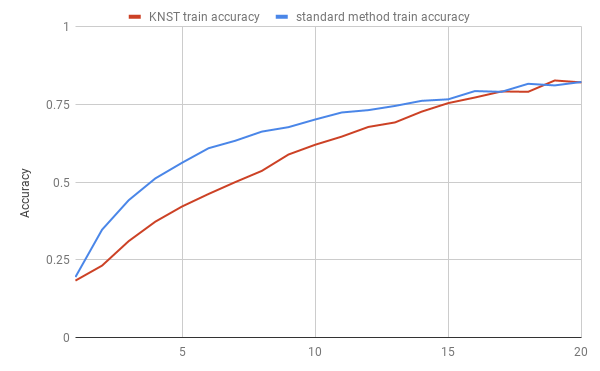}
	\caption{The training accuracy over the first 20 iterations for KNST and standard approach baseline.}
	\label{fig:knst_train_acc}
\end{figure}

\begin{figure}
	\includegraphics[width=\linewidth]{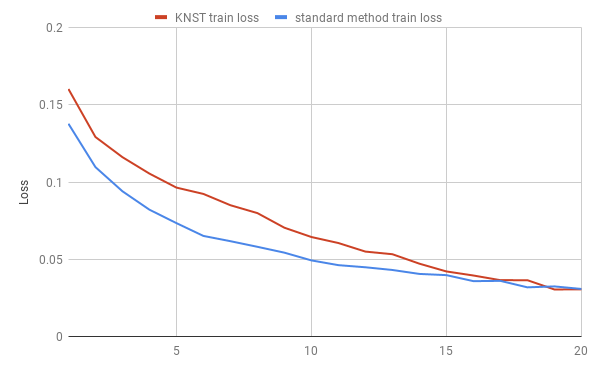}
	\caption{The training loss over the first 20 iterations for KNST and standard approach baseline}
	\label{fig:knst_train_loss}
\end{figure}

We demonstrate our results in Figure \ref{fig:knst_train_acc} with the training accuracy and Figure \ref{fig:knst_train_loss}, which includes the training loss over the first 20 iterations.
The training loss does not converge faster than the standard method and the training accuracy has no noticeable improvements. 
Further, we note that KNST even appears to converge somewhat slower than the standard method during the initial iterations.

These results demonstrate that there is no apparent improvement from the KNST method compared to the standard method.
Further, to stylize the first layer of the classifier model will require additional GPU memory, meaning that each training step for the KNST approach is longer than a single training step for the standard method.
However, we note that the final test accuracy of the KNST method was roughly two percent higher than the standard method test accuracy.
This indicates that while KNST may not have outperformed the standard method in terms of training time or accuracy, it may have lead to more general final features.

This experiment had several limitations.
We only used one image in the stylizer to stylizing the first kernel in classifier.
It is quite likely that this image cannot represent all features that appear in our dataset. 
Further, KNST essentially had to train two models simultaneously, which requires more time and resources to reach convergence for both models.
In addition, the loss calculated during the training process is backpropagated through both models; therefore, both stylizer and classifier are influenced by each other and the converging speed is slowed down.   
Hence, it's reasonable that KNST didn't perform well in training process acceleration. 

Based on these results, we have to admit that KNST is not a mature approach to accelerate the training process. 
However, this method could be a good consideration for alternative classification tasks. 
For example, if the game's visual aesthetics were significantly less like real world visual aesthetics.


\section{Decreasing Execution Time with Zero-shot and Pruning}
In the prior sections we demonstrated approaches that decreased training time and therefore computation power. 
However, if one wished to actually apply these approaches to derive game logs, this could also be a potentially lengthy process depending on hardware.
In the case of attempting to apply these methods to derive information on an ongoing livestream, these models would need to run at near real-time, even on machines with only a single GPU.

After training a CNN, the crucial problem is how to deploy this model and run it in real-time scenarios.
The required time to predict one input image or frame is dependent on the size of the model and the computation power and the number of available GPUs.
For example, ResNet152 has about 57.99 million convolutional parameters.
To go through our entire \textit{DOTA2} dataset, ResNet152 takes approximately 108 seconds or approximately 0.012 seconds per frame (with Ubuntu18.04, GTX1080, CUDA9.0, cudnn7) \footnote{This data is not recorded under strict conditions, the time varies among different hardware setups, software configurations, and execution setups.}. 
This is just fast enough to process livestreams in real-time, but requires high-end hardware.
Even so, there are certain non-relative and unused weights and layers in our trained CNN, and they occupy the limited GPU memory and slow the execution time. 
If we could find ways to crop out those unused weights and non-relative layers, we would expect a significant improvement to the CNN execution speed and reduction of GPU memory usage. 
Therefore in this section we focus on two approaches to decrease the run time of this model, zero-shot learning and network pruning.
In both cases we focus on shrinking or making use of less of the total neural network.

\subsection{Modified Zero-Shot Approach}
\label{sec: mzsa}
Zero-shot learning represents a wide set of approaches for deriving classifiers without training or at least without traditionally training on any data \cite{romera2015embarrassingly}. 
\citeauthor{chao2016empirical} \cite{chao2016empirical} describes an approach to derive a classifier from a trained neural network by investigating the activations of only a few layers.
Essentially the approach looks at the activations over each class, and uses the average activation for each class as a nearest-neighbor classifier, matching a new candidate image to the class with the most similar average activation.
This has the benefit of lowering training time, assuming one has a pre-trained neural network.
However here we are most interested in its ability to lower execution time, given that we only need to process a subset of our original model.

We chose to check the activations of the first convolutional layers.
The motivation is that the very first convolutional layers handle general features of the input, and these features typically remain the key to predict the final output.
This follows from the fact that we want to minimize runtime and therefore utilize the smallest possible set of layers.
This process markedly decreases the GPU usage since it excludes all other layers other than the first few layers.
Additionally, the execution speed is accelerated significantly because each neural network forwarding step only goes through these layers instead of full CNN.
In our experiment, we employed an ImageNet pre-trained ResNet152 model and made use of first \textbf{4 layers} of this model to build a classifier.
These four layers consist of one 7x7x64 convolutional layer, one 64 batch normalization layer, one ReLU activation layer, and one max-pool layer. 
We anticipated this approach could reduce the execution time of ResNet152 while still retaining reasonable accuracy. 

The exact procedure to apply this zero-shot approach is following, assuming access to a ResNet152 model pre-trained on ImageNet.
First, we prepare another model with only the first \textbf{4} layers of the ResNet152 model, and we copied the weights from the pre-trained model to this new model, which we call the zero-shot classifier model.
We then feed the \textit{DOTA2} dataset frames and corresponding labels to the zero-shot classifier model to create a $CxNxN$ ($C$ is the size of final layer out channel number, and $N$ is the kernel size) feature matrix for each category in the \textit{DOTA2} dataset.
Those feature matrices are the \textit{labels} that we used to classify the input frames.
We made use of \textit{Euclidean Distance} method, the square differences between the output and our collected label, as the standard to perform the prediction, which we choose the minimal dist2 value label as the predicting label.
Finally, we evaluate the test accuracy on our \textit{DOTA2} dataset. 

\subsection{Modified Zero-Shot Approach Evaluation} 
In Section \ref{sec: mzsa} we demonstrate the motivation and structure of this modified zero-shot approach
In this section, we evaluate the execution time of this modified zero-shot approach on \textit{DOTA2} dataset thoroughly against the standard approach.
Ideally, this approach will improve in execution time without a significant loss in accuracy.

We include the average test accuracy in Table \ref{tab:zero-shot-comparison}. 
The result shows that our expectation were not met in terms of a non-significant loss of accuracy, with all variations performing with about half the accuracy of the complete ResNet152. 
We included two variations in our results beyond the ``Only First 4 Layers" version described in the previous section.
The first variation adds the layer1 module, which has a total of 13 layers, and the second adds both the layer1 and layer2 modules, for a total of 37 layers.
We note a considerable improvement in the 13-layer and 37-layer experiments compared to the first-4-layer experiment: with the 13-layer classifier reaching about 51\% test accuracy and 37-layer classifier reaching 56\% test accuracy. 
However, even though these two new classifiers outperform the first-4-layer classifier; the accuracy they reach is insufficient for any real world scenarios.
Despite this we did see our predicted improvement in speed, with the Time(s) value of Table \ref{tab:zero-shot-comparison} referring to the time it took to process the complete \textit{DOTA2} dataset, cutting our execution time to a third compared to the full ResNet152 model.
Across the variations we found out that the runtime does not increase significantly as additional layers are added.
We suggested that the reason of non-significant runtime speed increase is that the most costly portion of running a CNN model is extracting high-level features from the inputs, which is exactly the first layers in a CNN; also, some events only occupy in a tiny part of whole input frame, and the zero-shot classifier cannot actually distinguish between all occurring events.
However, it appears we would quickly plateau by increasing the size of the zero-shot classifier, without reaching the level of accuracy we'd need to deploy this approach in a real world environment.

\begin{table}
\caption{A comparison of different number layers in Zero-Shot Learning.}
\label{tab:zero-shot-comparison}
\centering
\begin{tabular}{ |l|c|c|c| } 
\hline
  & \makecell{Only First \\ 4 Layers} & \makecell{After Layer1 \\ Module}  & \makecell{After Layer2 \\Module} \\ 
\hline
 \makecell{Average \\ Accuracy} & 42.43\% & 50.99\% & 56.25\% \\
\hline
 Time (s) & 31.002 & 32.586 & 34.113 \\
\hline
\end{tabular}
\end{table}

In this section, we demonstrated that using this modified zero-shot approach can reduce the GPU memory usage and execution time while sacrificing prediction accuracy.
This tradeoff is too harsh to deploy in our target scenario, realtime e-sports streams on a single GPU computer. 
However, depending on the scenario, such a loss of accuracy may be acceptable due to the size of datasets that need to be processed, say an initial processing step in a longer pipeline.

\subsection{Network Pruning}
Neural network pruning represents a class of algorithms for removing less important neurons from a typically pre-trained neural network in order to retain similar performance but with smaller memory and faster processing speed \cite{reed1993pruning}. 
Given our goals for this section, it follows to apply this approach in order to reduce computation power requirements and to decrease the execution time of our \textit{DOTA2} event recognition model.
We employed a particular neural network pruning algorithm referred to as ThiNet \cite{luo2017thinet}, to shrink and prune the trained ResNet152 model.
Most neural network pruning approaches are focusing on redundant CNNs such as the VGG model, and ThiNet is the one of the few that fits ResNet. 
ThiNet is an efficient and unified neural network framework that simultaneously accelerates and compresses CNN models in both training and forwarding stages \cite{luo2017thinet}.
The ThiNet framework decreases the filters in each convolutional layer while retaining almost the same performance. 
In our experiment, we employ the forwarding stage of the ThiNet framework to compress the pre-trained ResNet152 and decrease the execution time.
We made use of the ThiNet30 version of ResNet152, ThiNet50 version of ResNet152, and ThiNet70 version of ResNet152 and compared their structures and performance against the baseline model, ResNet152 in the standard method.

The differences across the different versions of ThiNet are marginal, focused on how many convolutional layers each ResNet module has. 
We demonstrate exact differences across each version in Table \ref{tab:thinet-structure}.
All four models share the same \textbf{conv1 layer}, \textbf{average pooling}, and final 10-class \textbf{fc layer} (representing the ten \textit{DOTA2} gameplay events), while the other four convolutional layer modules differ in the number of convolutional layers and number of filters. 
As Shown in Table \ref{tab:thinet-structure}, each cell in the convolutional layer modules represents the three inner kernel plane numbers in each residual block.
For example, ThiNet30 has 30\% of plane numbers as the ResNet152 in the first two planes in each residual block, with Thinet50 and ThiNet70 have 50\% and 70\% respectively. 
We showed that fewer inner planes imply fewer parameters, and we include the parameter number of each CNN model in Table \ref{tab:thinet-compare}.

\begin{table*}
\caption{A comparison of ThiNets and ResNet152 structure.}
\label{tab:thinet-structure}
\begin{tabular}{|c|c|c|c|c|}
\hline
Layer Name                & ThiNet30             & ThiNet50              & ThiNet70              & ResNet152             \\ \hline
conv1                     & \multicolumn{4}{c|}{7x7, 64, stride 2}                                                       \\ \hline
\multirow{3}{*}{con2\_x}  & \multicolumn{4}{c|}{3x3 maxpool, stride2}                                                    \\ \cline{2-5} 
                          & 19, 19, 64           & 32, 32, 64            & 44, 44, 64            & 64, 64, 64            \\ \cline{2-5} 
                          & [19, 19, 256] x 2    & [32, 32, 256] x 2     & [44, 44, 256] x 2     & [64, 64, 256] x 2     \\ \hline
\multirow{2}{*}{conv3\_x} & 38, 38, 256          & 64, 64, 256           & 89, 89, 256           & 128, 128, 256         \\ \cline{2-5} 
                          & [38, 38, 512] x 7    & [64, 64, 512] x 7     & [89, 89, 512] x 7     & [128, 128, 512] x 7   \\ \hline
\multirow{2}{*}{conv4\_x} & 76, 76, 512          & 128, 128, 512         & 179, 179, 512         & 256, 256, 512         \\ \cline{2-5} 
                          & [76, 76, 1024] x 35  & [128, 128, 1024] x 35 & [179, 179, 1024] x 35 & [256, 256, 1024] x 35 \\ \hline
\multirow{2}{*}{conv5\_x} & 153, 153, 1024       & 256, 256, 1024        & 358, 358, 1024        & 512, 512, 1024        \\ \cline{2-5} 
                          & [153, 153, 2048] x 2 & [256, 256, 2048] x 2  & [358, 358, 2048] x 2  & [512, 512, 2048] x 2  \\ \hline
classifier layer          & \multicolumn{4}{c|}{average pool, 10-d fc, sofrmax}                                          \\ \hline
\end{tabular}
\end{table*}

In our implementation, we trained the ThiNet-30, ThiNet-50, ThiNet-70 directly on our \textit{DOTA2} dataset and used them as the student in our teacher-student approach.

\subsection{Network Pruning Evaluation}

\begin{table}
\caption{A comparison of different ThiNets and original ResNet152.}
\label{tab:thinet-compare}
\centering
\begin{tabular}{ |l|c|c|c|c| } 
\hline
& ThiNet30 & ThiNet50  & ThiNet70 & ResNet152 \\ 
\hline
\hline
\makecell{Parameters} & 6573806 & 10287296 & 14847686 & 57992384\\
\hline
\makecell{Memory \\ Usage (MB)} & 1405 & 1473 & 1541 & 2669 \\
\hline
\makecell{Parameter- \\Memory \\ Ratio} & 4678.9 & 6983.9 & 9635.1 & 21728.1 \\
\hline
\makecell{Execution  \\ Time (s)} & 23.665 & 23.564 & 23.984 & 24.516 \\
\hline
\makecell{Test  \\ Accuracy} & 99.7\% & 99.8\% & 99.8\% & 94.6\% \\
\hline
\end{tabular}
\end{table}

In this section, we evaluate the different ThiNet models and ResNet-152 in terms of the number of parameters, GPU Memory Usage, Execution time, and test accuracy.
Shown in Table \ref{tab:thinet-compare}, the GPU memory usage of each CNN increases when the parameters of each CNN increases. 
Also, as the number of CNN's parameters increase, the parameter-memory ratio also increases.
We also note that the execution time for all four models, this time over only the test frames of our dataset was almost the same.
To the best of our knowledge, this result is due to the optimization of cudnn and that forwarding is not the most costly process.
The initializing processes of different models also have almost the same computation cost.
Thus the individual per-frame processing time is lower, but there is some start-up time for each model, and this is largely unchanged. 

We show this approach can effectively reduce the parameters of CNNs but due to the optimization of low-level code compilation, the differences in execution time are not distinct. 
Compared to the modified zero-shot approach we proposed in the previous section, ThiNet is a better solution to reduce the GPU memory usage while keeping high prediction accuracy.
The cause is that, in our experiment, all ThiNet models have 152 layers and they can fit the dataset much better than the modified zero-shot approach. 
In addition, there are significant savings in terms of memory usage, dropping to nearly half of the complete ResNet152 model.


\section{Takeaways}
We present two different approaches in this paper for reducing the requirements to apply an accurate and efficient CNN classifier to \textit{DOTA2}: reducing training time and runtime/memory requirements. 

For decreasing training time, the superior approach regarding performance was teacher-student with a ResNet152 model pre-trained on ImageNet serving as the teacher. This approach demonstrates that a large and high-quality pre-trained model could help reduce the training time and conserve computational resources. 
Compared to standard backpropagation approach, we anticipate the teacher-student transfer learning method can have a significant impact in the field of saving computational power and training time for applying a CNN classifier to e-sport gameplay event prediction from video. 
Our second approach for decreasing training time was kernel neural style transfer. 
This innovative approach took longer to train compared to the standard method and cost more computational resources. 
However, we provided some evidence that this approach lead to a more general final model.
We anticipate this method can be valuable for games with more highly stylized aesthetics.

For decreasing the GPU memory requirement, the superior approach was neural network pruning, in our experiment, the ThiNet framework can effectively reduce the parameters of the CNN and therefore the GPU memory usage.
Comparing to the standard method, the ThiNet models contains fewer parameters and GPU memory usage, and we anticipate this approach can be impactful in terms of increasing the types of machines capable of running these models.
The other approach we demonstrated is the zero-shot approach. This approach is fast and easy to apply but comparably inaccurate in prediction.

The approaches to decrease the training time and the ones to reduces the execution time are independent from each other, and every single approach can be employed separately for different situations.
We acknowledge that some of these strategies for reducing the requirements to apply accurate and efficient CNN classifiers to e-sports games will impact the performance of the CNN regarding test accuracy and stability. 
Our results in this paper only indicate that both aspects can be viable enhancements to reduce the requirements of applying CNN models. 
Together, we note that these improvements, the teacher-student transfer method and the network pruning with ThiNet allow for a high-quality model that can run in real-time on typical gaming computers.


\section{Limitations and Future Work}

Our work proves that a combined solution of training acceleration and running optimization is one suitable solution to reduce the requirements to apply accurate CNNs to game event prediction from e-sport gameplay video.
However, the teacher-student approach requires a teacher model pre-trained on a large, high quality dataset. 
While there are many such pre-trained models available, they cannot cover all possible scenarios, and we identify constructing a high quality dataset and training an initial teacher model as future problems.
For example, consider attempting to apply this approach on a retro game with a pixelated aesthetic. 
One would expect, and we have found from initial tests, that real world datasets do not transfer nearly as well to this aesthetic as the cartoony, but high quality aesthetic of \textit{DOTA2}. 
Thus, to apply the teacher-student method, one would need an existing, high-quality dataset with a pixelated aesthetic. 
While there have been prior attempts at this process \cite{luo2018player}, they are not of sufficient quality for real world applications.
The kernel neural style transfer approach is not a successful approach, but it represents an innovative strategy, and may be helpful in extreme cases like this. In future work we hope to address these issues in order to make these approaches accessible to even wider audiences.
We acknowledge that some of the strategies employed in this paper are mature and known, and this work serves as a proof that these techniques can help decrease the barrier of applying information extraction methods on \textit{DOTA2} and similar MOBA games.

While we anticipate that these approaches should extend to games with semi-realistic aesthetics at least as close as \textit{DOTA2}, we acknowledge we would likely need to make changes to our approach in certain situations. 
For example, for a first-person shooting e-sports game such as \textit{Overwatch}, which does not have the same over-the-top view as \textit{DOTA2} and therefore has more events that may occur off-screen. 
However other e-sport games like \textit{Hearthstone} and \textit{Artifact}, which are card games with a good deal of hidden information, may still be successful domains for this approach given that all cards are still visible once played. 
Besides applying these techniques and similar ones for gaming event extraction, we also look forward to extending the usage to non-gaming events in the real world, like sports.

We acknowledge that a current limitation of this work is that we do not actually attempt to run any of these approaches in the context of our goal scenario. 
While we note that this was meant primarily as a motivating example for the kind of applications one might imagine in the future from these technologies, we still anticipate a need to apply these methods in the real world.
This also underscores a limitation in our focus primarily on hardware accessibility, issues related to financial and technical knowledge accessibility will certainly arise in a real world application.
We anticipate the ability to address the technical knowledge barriers with explainable AI and other human-computer interaction methods, which will open further avenues for future research.


\section{Conclusions}
Our work proposed a combined solution to reduce the requirements for extracting information from the e-sport \textit{DOTA2} gameplay videos: specifically, how to decrease CNN training time, execution time, and memory requirements.
Since all four approaches are independent, the performance of the combination of methods would likely be an assembly of their performance.
We investigate the ability and performance of utilizing teacher-student transfer learning on \textit{ImageNet} and kernel neural style transfer of \textit{DOTA2} to train a CNN faster. 
We further demonstrate the impact of zero-shot prediction and neural network pruning on CNN execution time and memory usage.
We evaluate these approaches against the standard backpropagation method.
Further, we present a corpus we developed for \textit{DOTA2} event recognition.
Our results demonstrate that both approaches can help reduce the requirement of applying complex CNN classifiers to e-sport games and a combined solution of these two approaches has high potential for minimizing technical barriers to the accessibility of these approaches.


\begin{acks}
This material is based upon work supported by the National Science Foundation under Grant No. IIS-1525967. Any opinions, findings, and conclusions or recommendations expressed in this material are those of the author(s) and do not necessarily reflect the views of the National Science Foundation.
\end{acks}


\bibliographystyle{ACM-Reference-Format}
\bibliography{main}

\end{document}